\documentclass[letterpaper, 10 pt, conference]{ieeeconf}  
\usepackage{graphics} 
\usepackage{epsfig} 
\usepackage{mathptmx} 
\usepackage{times} 
\usepackage{amsmath} 
\usepackage{amssymb}  
\usepackage{url}
\usepackage[applemac]{inputenc}
\usepackage{comment}
\usepackage{bm}
\usepackage[dvipdfmx]{color}
\usepackage[labelformat=simple]{subcaption}
\usepackage{adjustbox}
\usepackage{float}

\title{\LARGE \bf
  Combining Ontological Knowledge and Large Language Model for
  User-Friendly Service Robots
}

\author{Haru Nakajima and Jun Miura\\
Department of Computer Science and Engineering\\
Toyohashi University of Technology
}

\begin{document}
\maketitle
\thispagestyle{empty}
\pagestyle{empty}

\begin{abstract}

Lifestyle support through robotics is an increasingly promising field,
with expectations for robots to take over or assist with chores like
floor cleaning, table setting and clearing, and fetching items. The
growth of AI, particularly foundation models, such as large language
models (LLMs) and visual language models (VLMs), is significantly
shaping this sector. LLMs, by facilitating natural interactions and
providing vast general knowledge, are proving invaluable for robotic
tasks. This paper zeroes in on the benefits of LLMs for "bring-me"
tasks, where robots fetch specific items for users, often based on
vague instructions. Our previous efforts utilized an ontology extended
to handle environmental data to decipher such vagueness, but faced
limitations when unresolvable ambiguities required user intervention
for clarity.  Here, we enhance our approach by integrating LLMs for
providing additional commonsense knowledge, pairing it with
ontological data to mitigate the issue of hallucinations and reduce
the need for user queries, thus improving system usability.  We
present a system that merges these knowledge bases and assess its
efficacy on "bring-me" tasks, aiming to provide a more seamless and
efficient robotic assistance experience.
  
\end{abstract}

\section{Introduction}
\label{sec:intro}

Lifestyle support is one of the promising application areas of
robotics technologies, and many chore tasks are expected to be
replaced or supported by service robots. Possible chores include floor
cleaning, setting and clearing tables, and finding and bringing
objects, among others, and various systems have been developed that
can interact with the users using knowledge about the task and
environment \cite{DeitsJHRI-2012,LilianaRAS-2021,ShinHRI23-Companion}.
Along with the rapid advancement of AI technologies, the expectations
for such service robots are greatly rising. The biggest recent
advancement is undoubtedly generative AI, especially those based on
foundation models, such as large language models (LLMs)
\cite{GPT-4,LLaMA2} and visual language models (VLMs) \cite{CLIP}.

Introducing LLMs into service robots is effective as they not only
enable natural dialog between robot and human but also provide general
knowledge that can be used for robot control
\cite{AhnArXiv-2022,HuangICML-2022,ProgPromptICRA-2023}. This paper
focuses on the latter advantage of LLMs. We deal with {\em bring-me}
tasks, in which a robot brings a user-specified object from a distant
place. Users' commands can be ambiguous, where complete information for
robot action execution is partially missing. In our previous work
\cite{LilianaRAS-2021}, we used the ontology with extension for
handling facts in the environment to resolve ambiguities. However, the
recorded facts are limited and the robot always asked the user for
necessary detailed information when the robot cannot resolve
ambiguities completely by itself.

In this paper, we adopt a large language model (LLM) as another source
of knowledge. In addition to being able to obtain commonsense-like
knowledge from the LLM, combining two knowledge sources can reduce the
hallucination problem by filtering outputs from the LLM using the
ontological knowledge. This is effective to reduce the number of
inquiries to the user, thereby increasing the usability of the
system. This paper develops a system that combines the two knowledge
sources and evaluates the performance using {\em bring-me} tasks.  One
of the crucial issues in applying LLMs to service robots is how to
prevent hallicinations, as we cannot completely address it even if we
tune prompts as much as possible. One approach is to introduce an
explicit mechanism for grounding outputs. Recent trends in
retrieval-augmented text generation (RAG) \cite{RAG-1-2021,RAG-2-2022}
seek similar directions. Some work has proposed to use knowledge or
experiences to generate feasible robotic plans
\cite{LuCLAP-2022,DingIROS-2023,SakimArXiv-2023,RAP-2024}. In this
paper, we take this approach by combining ontological knowledge with
LLMs. The contribution of the paper is twofold. First, we propose a
new method of integrating two knowledge sources. Second, we evaluate
the method for the ``bring-me'' task in terms of feasibility,
efficiency, and usability.

%

\section{Related Work}
\label{sec:RelatedWork}

\subsection{Service robots utilizing knowledge about the environment
  and dialog}

Home service robots must have knowledge about the working environment
and utilize it to automatically generate action plans. In
``bring-me'' tasks, for example, the type and the location of the
objects are crucial information. Several studies have proposed the use
of ontologies as basis of knowledge representation
\cite{OlszewskaROMAN-2017,BeblerAAMAS-2018,RicardezIRC-2018}.  We have
developed a robotic system that utilizes ontological knowledge
supported by verbal interaction to resolve ambiguities in users'
commands \cite{LilianaRAS-2021}.  When knowledge bases are incomplete
and uncertain, the robot may fail to retrieve correct or complete
information for task execution and needs to ask the user for the
missing information. Such a dialog-supported strategy is effective but
might degrade the usability of the system by repeated inquiries.

Generating appropriate queries for increasing the possibility of
obtaining informative answers is a key to developing user-friendly
service robots.  Morohashi and Miura \cite{MorohashiECMR-2019}
developed a method to generate queries that will yield the most
informative answer from the user. Pramanick et
al. \cite{PramanickRAS-2022} developed a method for ambiguity
resolution by identifying the type of ambiguity using visual scene
analysis and by generating appropriate inquiries for that type.  Shin
et al. \cite{ShinHRI23-Companion} proposed a framework enabling social
robots to generate questions that resolve uncertainties, taking into
account knowledge categories, required cognitive processes, and types
of questions. Generating appropriate queries will increase the
usability of the system, but it would be nice if we could reduce the
number of queries further.

\subsection{Large language models for robots}

Large language models (LLMs) have been more and more popular in
robotics, in addition to many other domains, and there are many
attempts to make robotic plans using LLMs
\cite{AhnArXiv-2022,HuangICML-2022,ProgPromptICRA-2023,DingIROS-2023,CodeAsPoliciesArXiv-2022,GramopadhyeArXiv-2022,OckerArXiv-2023,XieArXiv-2023}.
In these works, one of the common issues is how to make LLM outputs be
consistent with (or grounded on) the current situation. Many of such
works consider both the appropriateness and the feasibility of plans to
choose the best plan. Example command-plan pairs, which are
carefully-chosen or actually-held, are often input as prompts.  They
do not introduce a mechanism of explicitly filtering out ungrounded
information.

Ocker et al. \cite{OckerArXiv-2023} explored the use of LLMs as
common-sense knowledge base for the robot to understand the context of
household tasks. They showed potential applicability of LLMs but
suggested that combined use of LLMs and other knowledge sources would
be effective.  Lu et al. \cite{LuCLAP-2022} developed a mechanism for
extracting causal knowledge from the database, which is similar to the
current planning problem, to include it within the next prompt.  Ding
et al. \cite{DingIROS-2023} developed a system that can generate
feasible and efficient task plans by symbolic and geometric inference
by LLM followed by feasibility checking. LLM-GROP enables the robot to
achieve tasks with underspecified requests using LLM while
guaranteeing feasibility using a separate checking module. Sakim et
al. \cite{SakimArXiv-2023} proposed to use a knowledge network
\cite{Foon} for similar purpose.



%
%
%

\section{Overview of the Proposed Approach}
\label{sec:Overview}

\subsection{System configuration}
\label{subsec:SystemConfiguration}

\begin{figure}[tb]
\begin{center}
   \includegraphics[width=\linewidth]{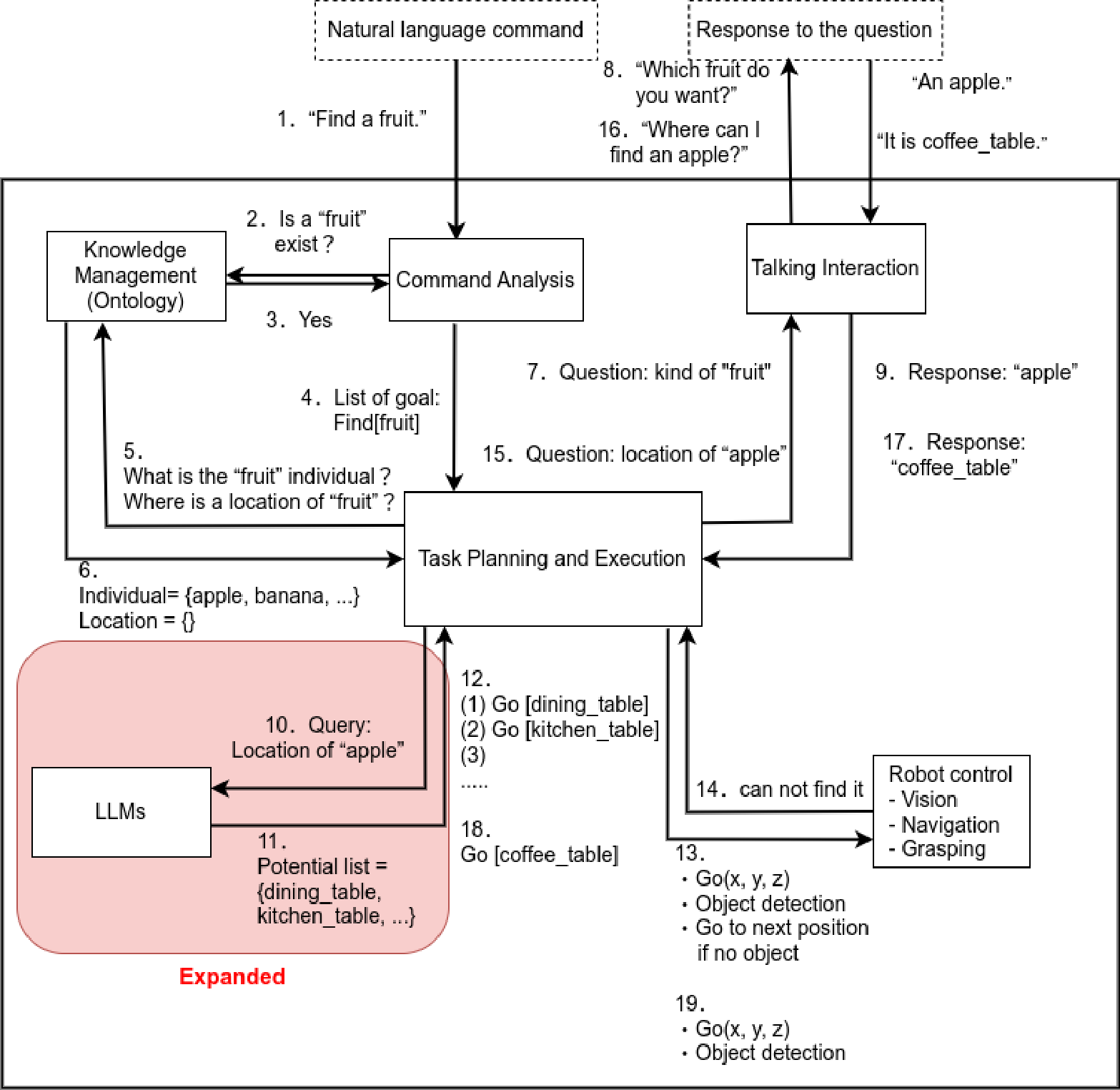}
   \caption{System overview.}
   \label{fig:Overview}
\end{center}
\end{figure}

Fig. \ref{fig:Overview} depicts the overview of the system
structure. The system is based on our previous system which utilizes
ontological knowledge and is extended to have a large language model
(LLM) as another knowledge source. We use Llama 2 \cite{LLaMA2} as the LLM.

The figure also shows an example interaction between the user and the
system. In the command ``Find a fruit,'' the name of the fruit and its
location are ambiguous. As preferences of fruits vary from person to
person, the system directly asks the user for their preference (steps
7 to 9). On the other hand, the location of ``apple'' may belong to
general knowledge (i.e., LLM), and the system refers to the LLM to
obtain a list of possible locations and generates a robot plan for
visiting these locations (steps 11 to 13). If the robot cannot find an
apple at these locations, the system can pose another inquiry to
complete the task (steps 15 to 19).

\subsection{Addressing Hallucination}
\label{subsec:AddressingHallucination}

LLMs sometimes suffer from hallucinations, which results in generating
incorrect outputs, or outputs inapplicable to the problem under
consideration. In this work, the robot is intended to work in ordinary
house environments, and we use an LLM to obtain information about
object locations. We address the hallucination problem by (1) giving as a
prompt a set of concrete knowledge of the environment, which is retreived
from our ontological knowledge base, and (2) verifying the consistency
between LLM outputs with the knowledge base. We will describe these
steps in detail in the following sections.

\section{Proposed System}
\label{sec:ProposedSystem}

\subsection{LangChain}

LangChain is a framework for developing applications powered by
language models, making the developed system context-aware and capable
of reasoning using the model \cite{LangChain}. Here, we describe the
use of LangChain in implementing our system.

\subsubsection{Memory}

Memory is used for incorporating history of dialog as contexts, by
keeping the history and giving it along with inquiries in prompts. This
enables the adoption of ``Chain-of-Thought'' reasoning
\cite{CoTArXiv-2022}.  We use ConversationTokenBufferMemory
functionality, which automatically includes the history with a token
size limitation. We limit the size of tokens such that the total text
size does not exceed the maximum size allowed in Llama 2. We use the
following expression as the token size limitation:
\begin{eqnarray}
  max\_token\_size = \left(max\_seq\_len / 2 -
  sys\_token\right)\times 0.8,\nonumber
\end{eqnarray}
where $max\_seq\_len$ is the maximum allowed token size of Llama 2 and $sys\_token$ is
the token size of the system prompt (see Sec. \ref{subsec:Prompting}).
This expression is for avoiding generating incomplete outputs for a
large $max\_seq\_len$ and for using 80\% of the total token size for
the context, so that Llama 2 can complete outputs even when the size
of inquiry is large. 

\subsubsection{Output parser}
\label{subsubsec:OutputParser}

\begin{figure}[tb]
\begin{center}
   \includegraphics[width=\linewidth]{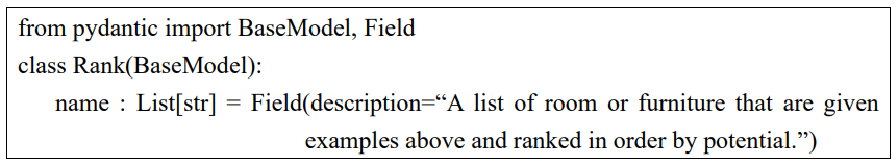}
   \caption{Class definition of output format.}
   \label{fig:OutputParserClass}
\end{center}
%
\medskip
\begin{center}
   \includegraphics[width=\linewidth]{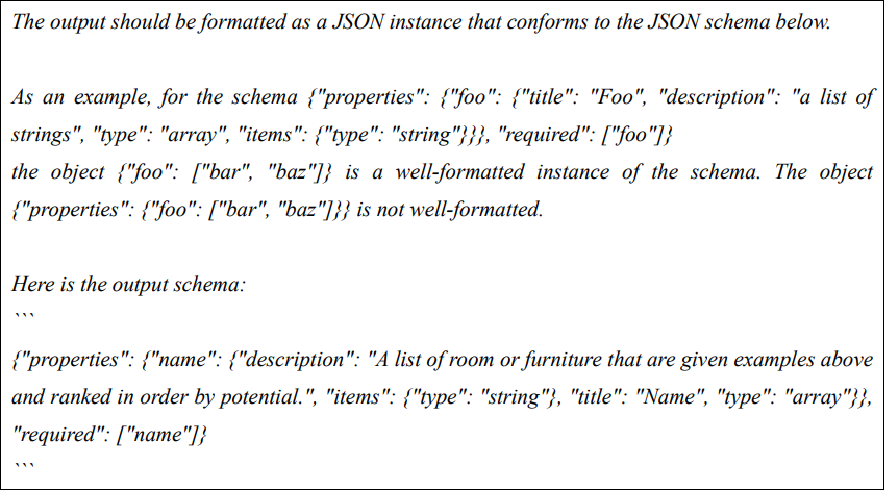}
   \caption{Embedding of the output format.}
   \label{fig:OutputParser}
\end{center}
\end{figure}

Output parser is the functionality to define the output format. We use
PydanticOutputParser and specify a JSON format.
Fig. \ref{fig:OutputParserClass} is the definition of the output
format class, indicating an output is a Python list of names of rooms
or furniture. Fig. \ref{fig:OutputParser} shows the definition of the
output JSON format.

\subsection{Prompting}
\label{subsec:Prompting}

\subsubsection{System prompt}
\label{subsubsec:SystemPrompt}

\begin{figure}[tb]
\begin{center}
   \includegraphics[width=\linewidth]{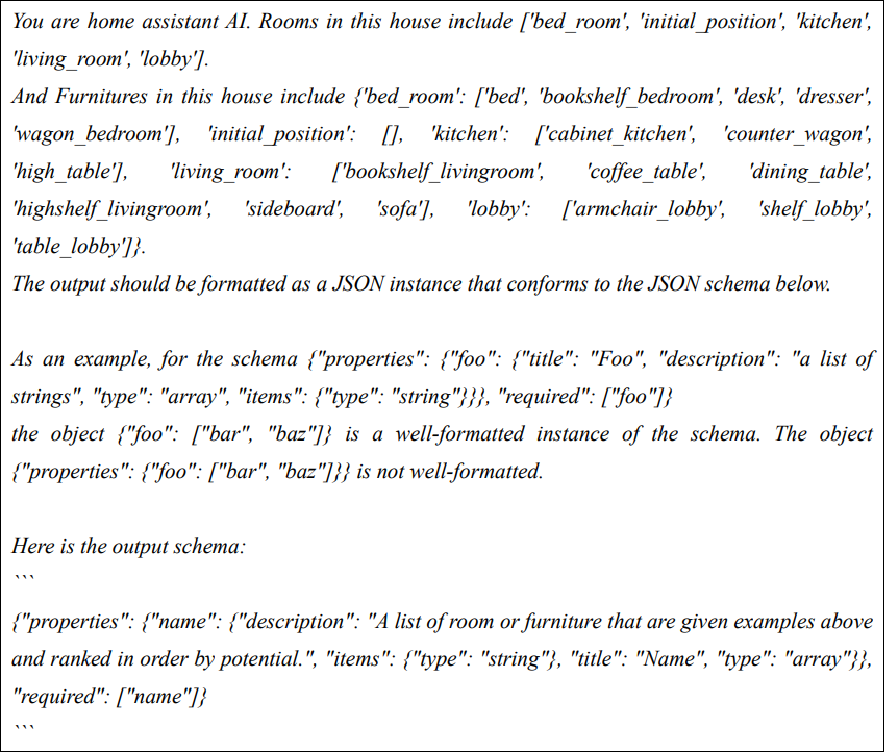}
   \caption{System prompt used in the experiment.}
   \label{fig:SystemPrompt}
\end{center}
\end{figure}

We embed two types of information as prompts. One is the output format
described in Sec. \ref{subsubsec:OutputParser}. The other is a set of
concrete knowledge of the environment obtained from the ontological
knowledge base. The latter includes the list of all rooms and
furniture in the environment and the furniture-room relationships
(i.e., in which room each furniture item exists). Giving such
information contributes to avoiding hallucinations about room or
furniture names. Additionally, giving it in the system prompt enables
the system to refrain from providing it repeatedly, thereby reducing
the size of tokens. Fig. \ref{fig:SystemPrompt} shows the system
prompt used in our system.

\subsubsection{Inquiry prompt}
\label{subsubsec:InquiryPrompt}

Table \ref{tab:InquiryPrompt} shows the template of the prompt used
for inquiry. Labels {\em general\_pos} and {\em multiple\_pos} are used for direct
inquiries from the robot. Label {\em general\_pos} is used when no default
locations are recorded for an object in the ontological knowledge
base. Label {\em multiple\_pos} is used when an object has multiple default
locations to constrain the output to such locations.

Labels {\em furniture}, {\em again}, and {\em summarize} are used for the consistency
check step. Label {\em furniture} is for obtaining a probable set of
furniture from the candidates. Label {\em again} is to repeat the inquiry in
case Llama 2 fails to output the location list. Label {\em summarize} is
used in case the output format is not appropriate. 

\begin{table}[tb]
\begin{center}
   \caption{Template for inquiry prompt.}
   \label{tab:InquiryPrompt}
   \begin{tabular}{|c|p{6cm}|}
     \hline
     Label & Inquiry\\
     \hline\hline
     {\em general\_pos} & I am searching for \#object name. Which places are most probably where I
     can find it? Please tell me the top 5.\\
     \hline
     {\em multiple\_pos} & I am searching for \#object name. Which places are most probably where I
     can find it, \#position? Please sort them in the order of
     likelihood.\\
     \hline
     {\em furniture} & I am searching for \#object name. Which furniture are most probably where
     I can find it in \#room name? Please tell me the top 3 furniture from
     furniture list, \#furniture list.\\
     \hline
     {\em again} & You must choose potential places from the list given
     above.\\
     \hline
     {\em summarize} & Please summarize about potential \#situation for \#object name. You must
     follow the given output format above.\\
     \hline
\end{tabular}
\end{center}
\end{table}

\begin{figure*}[t]
\begin{center}
   \includegraphics[width=\linewidth]{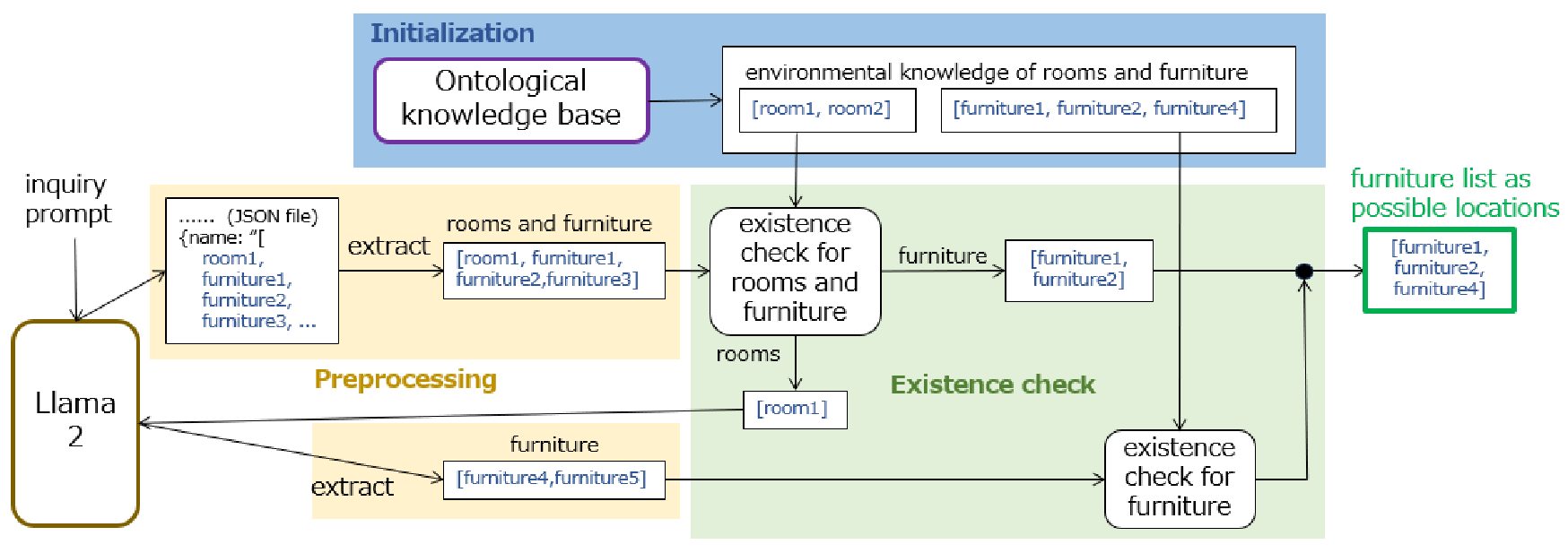}
   \caption{Consistency verification.}
   \label{fig:ConsistencyVerification}
\end{center}
\end{figure*}

\subsection{Consistency verification}
\label{subsec:ConsistencyVerification}

Consistency verification is responsible for ensuring the appropriateness of the
output, namely, the furniture list as possible locations of the target
object. The verification is composed of the following three parts
(see Fig. \ref{fig:ConsistencyVerification}):
\begin{itemize}
\item
  {\bf Initialization} part retrieves the concrete list of rooms and
  furniture from the ontological knowledge base.
\item
  {\bf Preprocessing} part extracts only possible locations from the
  JSON outputs from the Llama 2. When the output is not the JSON
  format\footnote{This could sometimes happen even though we give the
    prompt to output in the JSON format.}, it orders Llama 2 to
  re-generate the output using {\em summarize} label (see
  Sec. \ref{subsubsec:InquiryPrompt}). Then, it identifies the names
  of locations, which are either rooms or furniture, and sends them to
  the next part.
\item
  {\bf Existence check} part examines the existence of elements in the generated
  list with the retrieved list to make a list of existing names of
  furniture for possible locations. If there is a room name in the
  generated list, this part orders Llama 2 to make a list of furniture
  in the room, and adds them to the generated furniture list. The
  final output is the list of furniture as possible object locations. If
  the list is empty, this part returns the {\em not\_found} label, and
  the system resolves the ambiguity through dialog with the user.
\end{itemize}

\subsection{Integrated use of ontological knowledge base and LLM}

We develop the proposed system by extending the previous system
\cite{LilianaRAS-2021} that uses an ontological knowledge base.  The
ontological knowledge base is constructed by adding concrete
information about the environment to general ontology. On the other hand,
LLMs are based on large language datasets and thus cover a wide variety
of situations but may include information inapplicable to the current
environment. We thus prioritize the ontological knowledge base over
LLMs. Another point to note is that the system resolves the ambiguity
about objects only with either reference to the ontological knowledge
base or dialog with the user (as in the case of inquiry about
``fruit'' in Fig. \ref{fig:Overview}), because object preference is
personal in general and cannot be determined by LLMs' general
knowledge.

\begin{figure}[tb]
\begin{center}
   \includegraphics[width=0.8\linewidth]{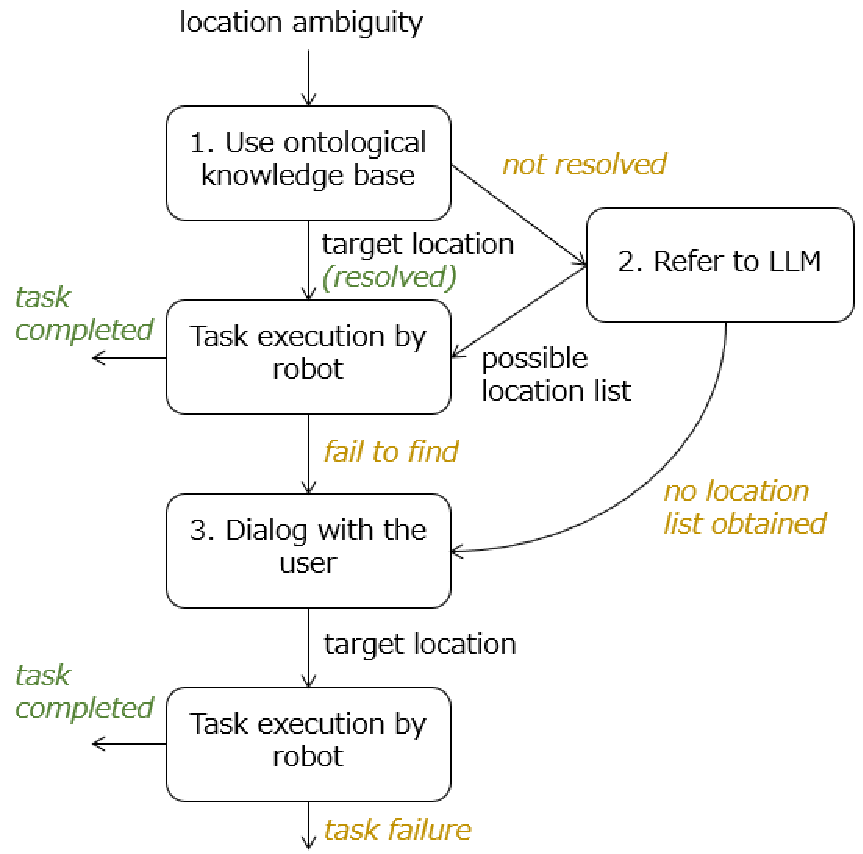}
   \caption{Flow of location ambiguity resolution.}
   \label{fig:ResolvingLocationAmbiguity}
\end{center}
\end{figure}

Fig. \ref{fig:ResolvingLocationAmbiguity} shows the flow of resolving
the ambiguity about object location. The system first consults with
the ontological knowledge base. If resolved, the system sends the
location to the robot to find the target object. If the robot fails to
find the object, the system asks the user about the location to
search. If the system cannot get a single location from the
ontological knowledge base, it then asks the LLM and, after the
consistency verification, sends the candidate location list to the
robot to visit. If the LLM does not return a location list, the system
asks the user.

\begin{figure*}[t]
\begin{center}
   \includegraphics[width=\linewidth]{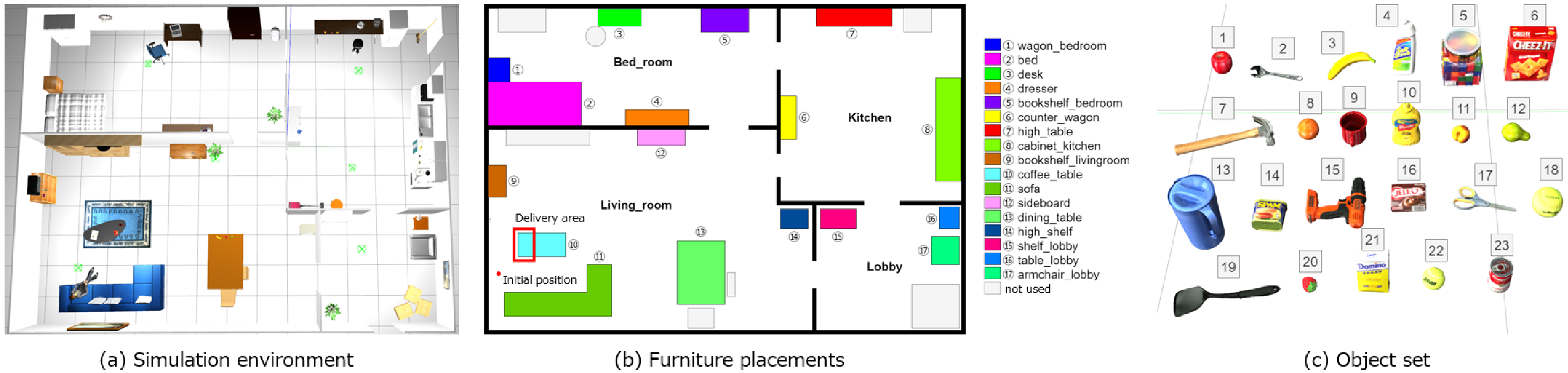}
   \caption{Testing environment and objects.}
   \label{fig:ExperimentalSetup}
\end{center}
\end{figure*}

\section{Experiments}
\label{sec:Experiments}

\subsection{Llama 2 implementation}

We use the Llama2-13b-chat model on a PC with Intel Xeon Gold 5222 and
Quadro RTX 800 on Ubuntu18.04 via Docker. The parameters used include:
max\_seq\_len=4096, max\_batch\_size=1, max\_gen\_len=2048,
temperature=0.6, and top\_p=0.9. We developed a ROS1 (Melodic)
interface to Llama 2 using ROS service. The service node on the Llama 2
server analyzes requests from the client and generates tokens for Llama
2. It also interprets the output from Llama 2 and sends the response
to the client.

\begin{figure}[tb]
\begin{center}
   \includegraphics[width=\linewidth]{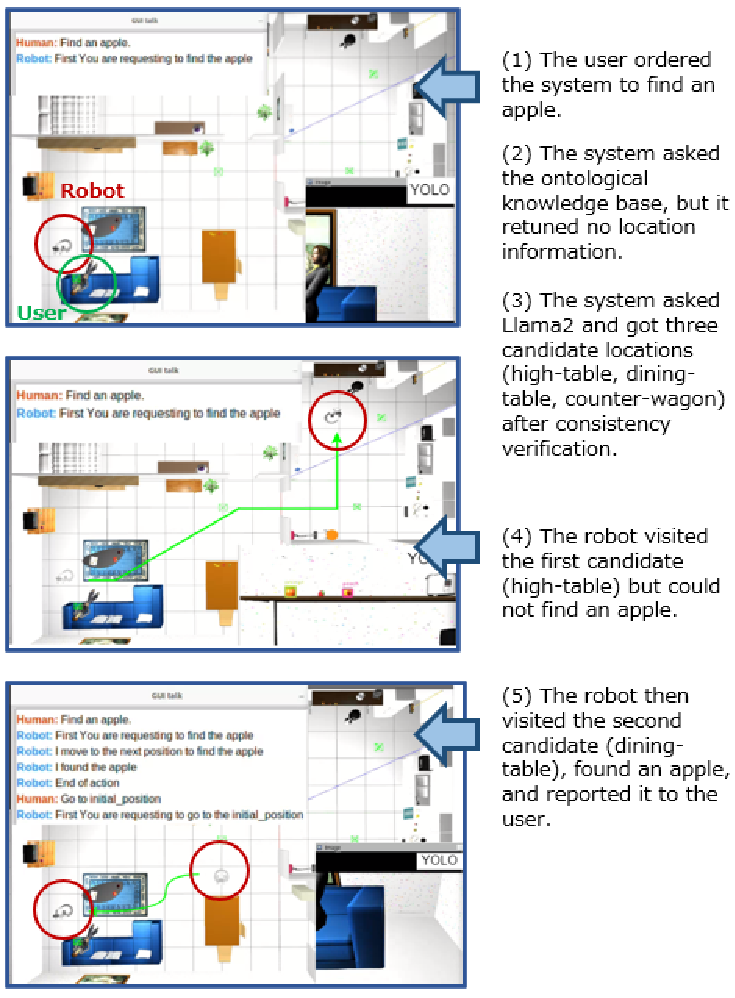}
   \caption{Example run.}
   \label{fig:ExampleRun}
\end{center}
\end{figure}

\begin{table}[tb]
\begin{center}
   \caption{Object set.}
   \label{tab:Objects}
   \begin{tabular}{|c|c||c|c|}
     \hline
     No. & Name & No. & Name \\
     \hline\hline
     1 & apple & 13 & pitcher\_base\\
     \hline
     2 & adjustable\_wrench & 14 & potted\_meat\_can\\
     \hline
     3 & banana & 15 & power\_drill\\
     \hline
     4 & bleach\_cleanser & 15 & power\_drill\\
     \hline
     5 & colored\_wood\_blocks & 16 & pudding\_box\\
     \hline
     6 & cracker\_box & 17 & scissors\\
     \hline
     7 & hammer & 19 & spatula\\
     \hline
     8 & orange & 20 & strawberry\\
     \hline
     9 & mug & 21 & sugar\_box\\
     \hline
     10 & mustard\_bottle & 22 & tennis\_ball\\
     \hline
     11 & peach & 23 & tomato\_soup\_can\\
     \hline
     12 & pear& &\\
     \hline
\end{tabular}
\end{center}
\end{table}

\subsection{Experimental settings}

We set up a four-room home environment (see
Fig. \ref{fig:ExperimentalSetup}(a)) using the Gazebo simulator. The
placements of furniture are shown in
Fig. \ref{fig:ExperimentalSetup}(b). The red point is the initial
robot position and the red rectangle is the delivery location in the
case that the user's order is to bring an object.
Fig. \ref{fig:ExperimentalSetup}(c) and Table \ref{tab:Objects} show
the set of objects used, which is a part of the YCB object set \cite{YCB}.
The robot uses YOLOv7 \cite{YOLO7} to recognize objects.
We added information about all the furniture, rooms, and objects to the ontology.
Additionally, each piece of furniture has an attribute indicating the room it is located in.

Fig. \ref{fig:ExampleRun} shows an example run. The user ordered the
system to find an apple. As the system could not find any location
information in the ontological knowledge base, it asked Llama 2 for
location information. After the consistency verification of the Llama
2's response, the system made a robot plan for visiting the candidate
locations one after another. The robot found an apple at the second
candidate (dining\_table) and came back to the user to report.

\subsection{Experimental conditions and metrics}

\begin{figure}[tb]
\begin{center}
   \includegraphics[width=\linewidth]{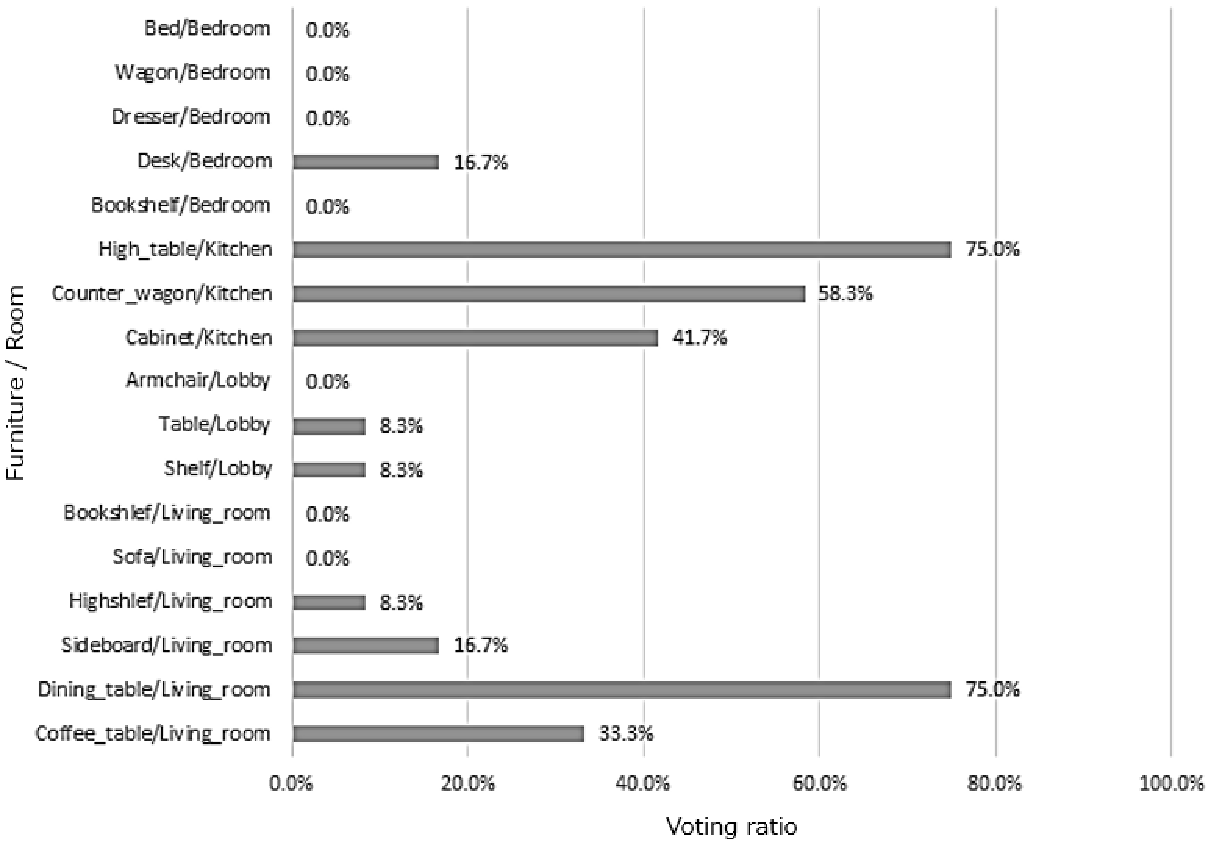}
   \caption{Example histogram for fruits.}
   \label{fig:FruitDistribution}
\end{center}
\end{figure}

We compare the following three approaches:
\begin{itemize}
\item
  (1) The system with only the ontological knowledge base (OKB)
  \cite{LilianaRAS-2021}. It uses OKB to get location information and
  asks the user when either no location information is available or
  multiple locations are obtained (denoted as OKB).
\item
  (2) The system with both OKB and LLM, where LLM does not use the
  dialog history (denoted as OKB+LLM). 
\item
  (3) The system with both OKB and LLM, where LLM uses the dialog
  history (denoted as OKB+LLM+MEM).
\end{itemize}
We also consider two situations: (a) OKB has default location
information, and (b) OKB does not have default location
information. Default location information is used for generating
location candidates in OKB, and for consistency verification in
approaches (2) and (3). We conducted a survey with sixteen subjects to
make the location distribution for each object, and extract high
probability locations as default
ones. Fig. \ref{fig:FruitDistribution} shows an example histogram for
``fruit'' class, and Table \ref{tab:DefaultLocations} shows the
complete list of object-location relationships. Some objects are
categorized into one class (e.g., fruits class).

We test all six combinations of the approaches and the situations
against a set of ten user commands shown in Table \ref{tab:Commands}.
We execute each command ten times.  We also simulate the user in the
experiments such that the user always gives correct answers when
asked.  The results are evaluated by the following metrics: task
completion rate, task completion time, number of user interaction, and
the number of furniture pieces visited.

\begin{table}[tb]
\begin{center}
   \caption{Default locations for each object class.}
   \label{tab:DefaultLocations}
   {\scriptsize
   \begin{tabular}{|c|c|}
     \hline
     Object & Default location list\\
     \hline\hline
     mug & [dining\_table, coffee\_table, counter\_wagon]\\
     \hline
     cracker\_box & [cabinet\_kitchen, dining\_table, coffee\_table, counter\_wagon]\\
     \hline
     fruits & [dining\_table, counter\_wagon, high\_table]\\
     \hline
     mustard\_bottle &[cabinet\_kitchen, dining\_table, counter\_wagon, high\_table]\\
     \hline
     spatula & [cabinet\_kitchen, counter\_wagon, high\_table]\\
     \hline
     power\_drill & [shelf\_lobby]\\
     \hline
     pitcher\_base & [cabinet\_kitchen, dining\_table, counter\_wagon, high\_table]\\
     \hline
     ball & [shelf\_lobby]\\
     \hline
     pudding\_box & [cabinet\_kitchen, dining\_table, coffee\_table, counter\_wagon]\\
     \hline
     colored\_wood\_blocks & (not available)\\
     \hline
     bleach\_cleanser & [cabinet\_kitchen, counter\_wagon]\\
     \hline
     potted\_meat\_can & [cabinet\_kitchen, counter\_wagon, high\_table]\\
     \hline
     tomato\_soup\_can & [cabinet\_kitchen, counter\_wagon, high\_table]\\
     \hline
     sugar\_box & [cabinet\_kitchen, counter\_wagon, high\_table]\\
     \hline
     hammer & [shelf\_lobby]\\
     \hline
     adjustable\_wrench & [shelf\_lobby, sideboard]\\
     \hline
     scissors & [sideboard, desk]\\
     \hline
   \end{tabular}
   }
\end{center}
\end{table}

\begin{table}[tb]
\begin{center}
   \caption{Set of commands for testing.}
   \label{tab:Commands}
   \begin{tabular}{|c|c|}
     \hline
     Command & Object location\\
     \hline\hline
     Find an apple. & dining\_table\\
     \hline
     Find a power\_drill. & shelf\_lobby\\
     \hline
     Find a colored\_wood\_blocks. & bookshelf\_bedroom\\
     \hline
     Find a pitcher. & counter\_wagon\\
     \hline
     Find a potted\_meat\_can. & cabinet\_kitchen\\
     \hline
     Take a peach. & high\_table\\
     \hline
     Take a mug. & coffee\_table\\
     \hline
     Bring a mustard\_bottle. & cabinet\_kitchen\\
     \hline
     Bring a sugar\_box. & counter\_wagon\\
     \hline
\end{tabular}
\end{center}
\end{table}

\subsection{Results}
\label{subsec:Results}

\subsubsection{Task completion rate}

Table \ref{tab:SuccessRate} summarizes the task completion rates for
each of the approaches. OKB and OKB+LLM+MEM exhibit remarkable
performance, while OKB+LLM was not competitive. Failure cases occur
due to object recognition failures. Adding context is obviously
effective in using LLM.

\begin{table}[tb]
\begin{center}
   \caption{Task completion rate}
   \label{tab:SuccessRate}
   \renewcommand{\arraystretch}{1.2}
   \begin{tabular}{|c|c|c|}
     \hline
     OKB & OKB+LLM & OKB+LLM+MEM\\
     \hline
     {\bf 98.0\%} & 80.0\% & \underline{97.0}\%\\
     \hline
\end{tabular}
\end{center}
\end{table}

\subsubsection{Time for task completion}

Table \ref{tab:TimeForTaskCompletion} compares the times for
completing tasks (only for success cases) for every combination of
approach, situation, and command type. The time increases as the
number of locations visited increases. OKB can achieve the task most
efficiently compared to the others. This is because in OKB, the system
asks the user every time there are any ambiguities and, therefore, can
get a single, correct target location, thereby minimizing the number of
visits. However, this merit must be considered with the demerit of
asking the user many times.

\begin{table}[tb]
\begin{center}
   \caption{Average task completion time [sec].}
   \label{tab:TimeForTaskCompletion}
   \renewcommand{\arraystretch}{1.2}
   \begin{tabular}{|c||c|c|c|c|c|c|}
     \hline
     & \multicolumn{3}{|c|}{without default location} & \multicolumn{3}{|c|}{with default location}\\
     \cline{2-7}
     & OKB &  OKB & OKB & OKB &  OKB & OKB \\
     &  &  +LLM & +LLM &  &  +LLM & +LLM \\
     &  &   & +MEM & &  & +MEM \\
     \hline\hline
     Find & \underline{83.0} & 311.8 & 338.3 & {\bf 81.8} & 143.8 & 157.4\\
     \hline
     Take & {\bf 99.3} & 369.9 & 395.4 & \underline{105.4} & 123.4 & 177.3\\
     \hline
     Bring & {\bf 224.7} & 301.7 & 304.4 & \underline{225.1} & 415.7 & 258.0\\
     \hline
\end{tabular}
\end{center}
\end{table}

\subsubsection{Number of inquiries to the user}

\begin{figure}[htb]
   \begin{center}
   \begin{minipage}{\linewidth}
    \begin{center}
    \begin{table}[H]
      \begin{center}
      \caption{Average number of inquiries to the user.}
      \label{tab:NumberOfInquiries}
      \begin{adjustbox}{width=\linewidth}
        \renewcommand{\arraystretch}{1.2}
        \begin{tabular}{|c||c|c|c||c|c|c|}
          \hline
          & \multicolumn{3}{|c||}{without default location} &      \multicolumn{3}{|c|}{with default location}\\
          \cline{2-7}
          & OKB &  OKB & OKB & OKB &  OKB & OKB \\
          &  &  +LLM & +LLM &  &  +LLM & +LLM \\
          &  &   & +MEM & &  & +MEM \\
          \hline\hline
          Mean & 1.0 & \underline{0.33} & {\bf 0.28} & 0.8 & \underline{0.14} & {\bf 0.08}\\
          \hline
          SD & {\bf 0.0} & 0.47 & \underline{0.45} & 0.40 & \underline{0.35} & {\bf 0.27}\\
          \hline
        \end{tabular}
      \end{adjustbox}
      \end{center}
    \end{table}
    \end{center}
  \end{minipage}
  \begin{minipage}{\linewidth}
    \begin{center}
    \begin{figure}[H]
      \begin{center}
      \includegraphics[width=\linewidth]{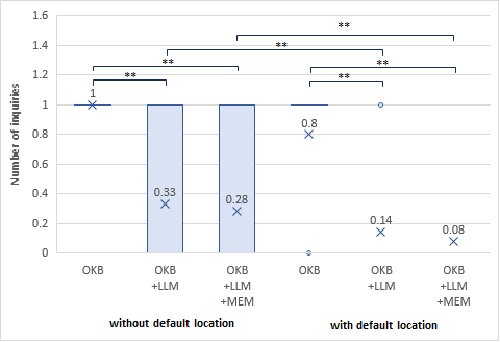}
      \caption{Statistical analysis of the number of inquiries to the user. 
              (*: \textit{p} \textless 0.05, **: \textit{p} \textless 0.01.)}
      \label{fig:NumOFInquiries}
      \end{center}
    \end{figure}
    \end{center}
  \end{minipage} 
  \end{center}
\end{figure}

Table \ref{tab:NumberOfInquiries} shows the number of inquiries to 
the user for each case, and Fig. \ref{fig:NumOFInquiries} shows the 
results indicating statistically significant differences. In these 
statistics, we assume that object recognition and robot action execution 
are always successful to focus on the capability of generating location 
candidates. From the table, where the results are analyzed separately for 
cases with and without default locations, it is evident that introducing 
an LLM significantly reduces the number of inquiries. This suggests a
considerable improvement in usability. In addition, using the dialog
history also has a positive effect.

\subsubsection{Number of visited pieces of furniture}

\begin{figure}[hbt]
  \begin{center}
  \begin{minipage}{\linewidth}
    \begin{center}
    \begin{table}[H]
      \begin{center}
      \caption{Average number of visited furniture.}
      \label{tab:NumberOfVisitedFurniture}
      \begin{adjustbox}{width=\linewidth}
        \renewcommand{\arraystretch}{1.2}
        \begin{tabular}{|c||c|c|c||c|c|c|}
          \hline
          & \multicolumn{3}{|c||}{without default location} &      \multicolumn{3}{|c|}{with default location}\\
          \cline{2-7}
          & OKB &  OKB & OKB & OKB &  OKB & OKB \\
          &  &  +LLM & +LLM &  &  +LLM & +LLM \\
          &  &   & +MEM & &  & +MEM \\
          \hline\hline
          Mean & {\bf 1.0} & \underline{3.68} & 3.70 & {\bf 1.0} & 2.45 & \underline{1.96}\\
          \hline
          SD & {\bf 0.0} & \underline{2.35} & 2.41 & {\bf 0.0} & 1.84 & \underline{1.45}\\
          \hline
        \end{tabular}
      \end{adjustbox}
      \end{center}
    \end{table}
    \end{center}
  \end{minipage}
  \begin{minipage}{\linewidth}
    \begin{center}
    \begin{figure}[H]
      \begin{center}
      \includegraphics[width=\linewidth]{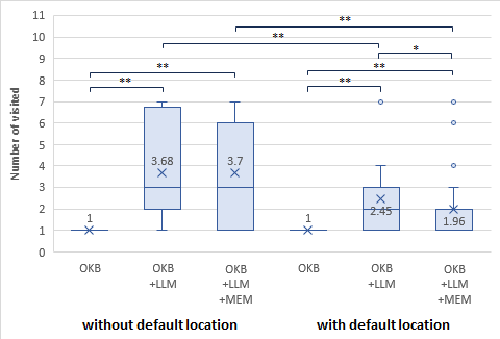}
      \caption{Statistical analysis of the number of visited furniture. 
              (*: \textit{p} \textless 0.05, **: \textit{p} \textless 0.01.)}
      \label{fig:NumOFVisitedFurniture}
      \end{center}
    \end{figure}
    \end{center}
  \end{minipage}
  \end{center}
\end{figure}

Table \ref{tab:NumberOfVisitedFurniture} shows the number of visited
pieces of furniture for each case, and Fig. \ref{fig:NumOFVisitedFurniture} 
shows the results indicating statistically significant differences. Here, we also assume that
recognition and action always succeed and have analyzed cases with and
without default locations separately. The OKB approach always visits
only one piece of furniture because the correct answer is obtained
either as ontological knowledge or from the user. In contrast, more
pieces of furniture are visited for the other cases (i.e., ones with
LLM). Among the approaches with LLM, using the dialog history is
effective. More interestingly, irrespective of using the dialog
history, the cases with default location information give better
results than the ones without it. This is because default location
information is utilized to effectively generate candidate location
lists with more promising furniture.

\subsubsection{Execution Cost}

Table \ref{tab:ExecutionCost} summarizes the average execution time
and count for each case. With default location information, the time
and the number of inquiries to LLM become shorter. Time for generating
outputs is almost the same for cases with and without default
location, while the approach using the dialog history needs extra time
for managing the memory, with generating more tokens.

\begin{table}[htb]
\begin{center}
   \caption{Average execution time and count.}
   \label{tab:ExecutionCost}
   \renewcommand{\arraystretch}{1.2}
   \begin{tabular}{|c||c|c|c|c|}
     \hline
     & \multicolumn{2}{|c|}{without} & \multicolumn{2}{|c|}{with}\\
     & \multicolumn{2}{|c|}{default location} & \multicolumn{2}{|c|}{default location}\\
     \cline{2-5}
     & OKB & OKB &  OKB & OKB \\
     &  +LLM & +LLM &  +LLM & +LLM \\
     &   & +MEM &  & +MEM \\
     \hline\hline
     Time for list generation [s] & 35.27 & 30.79 & \underline{10.81} & {\bf 9.95}\\
     \hline
     Number of inquiries to LLM & 4.63 & 3.14 & \underline{1.80} & {\bf 1.13}\\
     \hline
     Time for each LLM output [s] & {\bf 7.62} & 9.81 & \underline{7.51} & 8.75\\
     \hline
     Size of generated tokens & \underline{201.2} & 246.4 & {\bf 198.8} & 223.4\\
     \hline
\end{tabular}
\end{center}
\end{table}

\section{Summary}

\label{sec:Conclusion}

This paper has shown a new approach to combining ontological knowledge
with a large language model (LLM) for the ``bring-me'' task. The
developed system leverages the LLM, which has vast commonsense
knowledge about the household environment. As a result, it can
retrieve appropriate knowledge about object locations without overly
asking the user, thereby increasing usability. We developed a
framework to filter out infeasible solutions by repeatedly applying
firm knowledge in the ontological knowledge base. Although the
proposed approach seeks to prioritize generating feasible plans,
further improvements are needed for better handling the variety of
tasks as well as applying the proposed approach to real service robot
scenarios.

\section*{Acknowledgment}

This work is in part supported by the Kayamori Foundation of
Information Science Advancement.


\bibliographystyle{IEEEtran}
\bibliography{reference}

\end{document}